\documentclass[11pt,a4paper]{article}
\usepackage[hyperref]{acl2021}
\usepackage{times}
\usepackage{latexsym}
\usepackage{graphicx}
\usepackage{paralist}
\usepackage{multirow}
\usepackage{hyperref}
\usepackage[capitalise]{cleveref}

\usepackage{listings}
\lstdefinelanguage{json}{
  basicstyle=\ttfamily,
  columns=fullflexible,
  frame=single,
  breaklines=true,
}

\usepackage{color}

\usepackage{microtype}

\aclfinalcopy 

\title{PALI at SemEval-2021 Task 2: Fine-Tune XLM-RoBERTa for Word in Context Disambiguation}

\author{Shuyi Xie, Jian Ma, Haiqin Yang$^{\S}$,  Lianxin Jiang, Yang Mo, and Jianping Shen\\
  Ping An Life Insurance, Ltd. \\
  Shenzhen, Guangdong province, China \\
  {\small\{XIESHUYI542, MAJIAN446, JIANGLIANXIN769, MOYANG853, SHENJIANPING324\}@pingan.com.cn}\\
  $^{\S}$ {\small the corresponding author, email: {hqyang@ieee.org}}
}
\date{}

\begin{document}
\maketitle
\begin{abstract}
This paper presents the PALI team’s winning system for SemEval-2021 Task 2: Multilingual and Cross-lingual Word-in-Context Disambiguation.  We fine-tune XLM-RoBERTa model to solve the task of word in context disambiguation, i.e., to determine whether the target word in the two contexts contains the same meaning or not.  In implementation, we first specifically design an input tag to emphasize the target word in the contexts.  Second, we construct a new vector on the fine-tuned embeddings from XLM-RoBERTa and feed it to a  fully-connected network to output the probability of whether the target word in the context has the same meaning or not.  The new vector is attained by concatenating the embedding of the [CLS] token and the embeddings of the target word in the contexts.  In training, we explore several tricks, such as the Ranger optimizer, data augmentation, and adversarial training, to improve the model prediction.  Consequently, we attain the first place in all four cross-lingual tasks.
\end{abstract}

\section{Introduction}

This year, the SemEval-2021 task 2, multilingual and cross-lingual word-in-context (WiC)  disambiguation~\cite{martelli-etal-2021-mclwic}, defines the task of identifying the polysemous nature of words without relying on a fixed sense inventory in a multilingual and cross-lingual setting.  The task aims to perform a binary classification task to determine whether the target word contains the same meaning or not in two given contexts under both the same language (multilingual) setting and the different languages (cross-lingual) setting.  In the multilingual setting, the tasks consist of English-English (En-En), Arabic-Arabic (Ar-Ar), French-French (Fr-Fr), Chinese-Chinese (Zh-Zh) and Russian-Russian (Ru-Ru) while in the cross-lingual setting, the tasks consist of English-Chinese (En-Zh), English-French (En-Fr), English-Russian (En-Ru), and English-Arabic (En-Ar).  

The tasks contain the following challenges: 
\begin{compactitem}
\item The same word may deliver different meanings in different context~\cite{lei2021have}. 
\item The training data is scarce.  For example, in the multilingual tasks, there is only training data in En-En, while in the cross-lingual tasks, there is no training data.
\end{compactitem}

To overcome these challenges, we explore the uniqueness of the tasks and implement several key technologies: 
\begin{compactitem}
\item First, we follow~\cite{DBLP:conf/emnlp/BothaSG20} to specially design an input tag for the multilingual pre-training XLM-RoBERTa model to emphasize the target word in the contexts.  That is, the target word is encompassed by the special symbols of $<$t$>$ and $<$/t$>$.  Meanwhile, the given two contexts are concatenated by the $<$SEP$>$ token.  
\item Second, we apply data augmentation and add external data from WordNet to enrich the training data.  It is noted that we only expand the data in the task of En-En and do not consider other techniques, e.g., back-translation, for the cross-lingual tasks.  Adversarial training is also applied to learn more robust embeddings for target words.  The Ranger optimizer with the look-ahead mechanism in the AdamW optimizer is adopted to speed up the convergence of training.
\item Finally, we construct a new vector on the fune-tuned embeddings, i.e., concatenating the embeding of the [CLS] token and the learned embeddings of the target words' in both contexts.  The new vector is then fed into a fully-connected network to produce the binary classification prediction.  Cross-validation and model ensemble are also applied to attain a robust output.
\end{compactitem}
The rest of this paper is organized as follows: In Sec.~2, we briefly introduce related work.  In Sec.~3, we detail our proposed system.  In Sec.~4, we present the experimental setup, procedure, and the results.  Finally, we conclude our work in Sec.~5. 


\section{Related Work}
The SemEval-2021 task 2 aims to handling the tasks of  multilingual and cross-lingual word-in-context disambiguation~\cite{martelli-etal-2021-mclwic}, i.e., to determine whether the target word contains the same meaning in both given contexts.  In the following, we will elaborate several related work. 

Some recent effort, e.g.,~\cite{Pilehvar2018}, has been conducted to curate and release datasets to solve the task of WiC disambiguation.  Though it can be narrowed down to binary classification, some techniques have to be implemented to enhance the model performance.  For example, the trick of input highlighting mechanism~\cite{DBLP:conf/emnlp/BothaSG20} can be facilitated to promote the importance of the target word.  The idea of unifying entity linking and word sense disambiguation~\cite{DBLP:journals/tacl/0001RN14} can be borrowed to solve the task.  The idea of freezing the trained model for other languages~\cite{DBLP:conf/acl/ArtetxeRY20} can be explored to relieve the issue of no training data in the cross-lingual tasks.



Recently, due to the superior performance in tackling NLP tasks~\cite{conf/ijcai/PHED21,conf/ijcnn/EDM21,conf/ijcnn/RefBERT21}, pre-trained language models, such as BERT~\cite{DBLP:conf/naacl/DevlinCLT19} and RoBERTa~\cite{Liu2019}, start to dominate the way of word representations than static word embedding methods, e.g., Word2Vec~\cite{DBLP:journals/corr/abs-1301-3781} and FastText~\cite{DBLP:conf/eacl/GraveMJB17}.  Especially, the XLM-RoBERTa~\cite{DBLP:conf/acl/ConneauKGCWGGOZ20} model is a newly released large cross-lingual language model based on RoBERTa and is trained on 2.5TB filtered CommonCrawl data in 100 languages.  Different from other XLM models, XLM-RoBERTa does not require the language token to understand which language is used and can determine the correct language from the input ids.  It is a powerful tool for understanding multilingual languages and is very helpful for solving the WiC disambiguation task under the cross-lingual setting.  Hence, we choose XLM-RoBERTa in our system.


A critical issue of the task is lack of training data.  Though existing methods, e.g., lexical substitution~\cite{DBLP:conf/nips/ZhangZL15}, back translation~\cite{DBLP:conf/nips/XieDHL020}, and data augmentation~\cite{DBLP:journals/corr/FadaeeBM17}, can be applied to enrich the data, we mainly explore the usage of WordNet~\cite{wordnet} and the technique of pseudo labelling \cite{DBLP:journals/tip/WuP18} because WordNet contains rich synonyms while pseudo labelling is effective to utilize the abundant unlabeled data via their pseudo labels. 

Adversarial training~\cite{DBLP:conf/iclr/TramerKPGBM18} is an effective method to regularize parameters by introducing noise and to improve model robustness and generalization.  We also explore its possibility in fine-tuning XLM-RoBERTa to increase the robustness of the learned the word embeddings. 

\section{Overview}
In the following, we present the task definition, data preprocessing, and our proposed system design. 

\subsection{Task Defintion}
The task of WiC disambiguation is framed by a binary classification task.  Each instance in WiC has a target word $w$, whose part-of-speech is in \{NOUN, VERB, ADJ, ADV\}, with two given contexts, $c_1$ and $c_2$.  Each of these contexts triggers a specific meaning of $w$.  The task is to identify if the occurrences of $w$ in $c_1$ and $c_2$ correspond to the same meaning or not.  \Cref{fig:ex} illustrates an example from the dataset. 

\begin{figure}[htp]
\begin{center}
\begin{ttfamily}
\begin{tabular}{|@{~}l@{}p{6.1cm}|}
\hline
\{&\\
"ta&rget word": "play", \\
"se&ntence1": "In that context of coordination and integration, Bolivia holds a key play in any process of infrastructure development.", \\
"se&ntence2": "In schools, when water is needed, it is girls who are sent to fetch it, taking time away from their studies and play."\\
\}&\\
\hline
\end{tabular}
\end{ttfamily}
\end{center}
    \caption{An example from the WiC disamguation task.}
    \label{fig:ex}
\end{figure}

\subsection{Data Preprocessing}
\label{sec:length}

\begin{table}[htp]
    \centering
    \begin{tabular}{@{~}l@{~}|@{~}c@{~}|@{~}c@{~}}
    \hline
   & Training & Test\\\hline
  No. of target words & 3,726 & 491\\
  No. of pairs & 8,000 & 1,000\\
  Min. tokens & 6 & 5\\
  Avg. tokens (original) & 24 &26\\
  Max. tokens (original) & 88& 116\\
  Max. tokens (post-proc.) & 81 & 81 \\ 
  \hline
    \end{tabular}
    \caption{Statistics of the data\label{tab:data_stat}}
\end{table}
\begin{figure*}[htbp]
\centering
\includegraphics[scale=0.7]{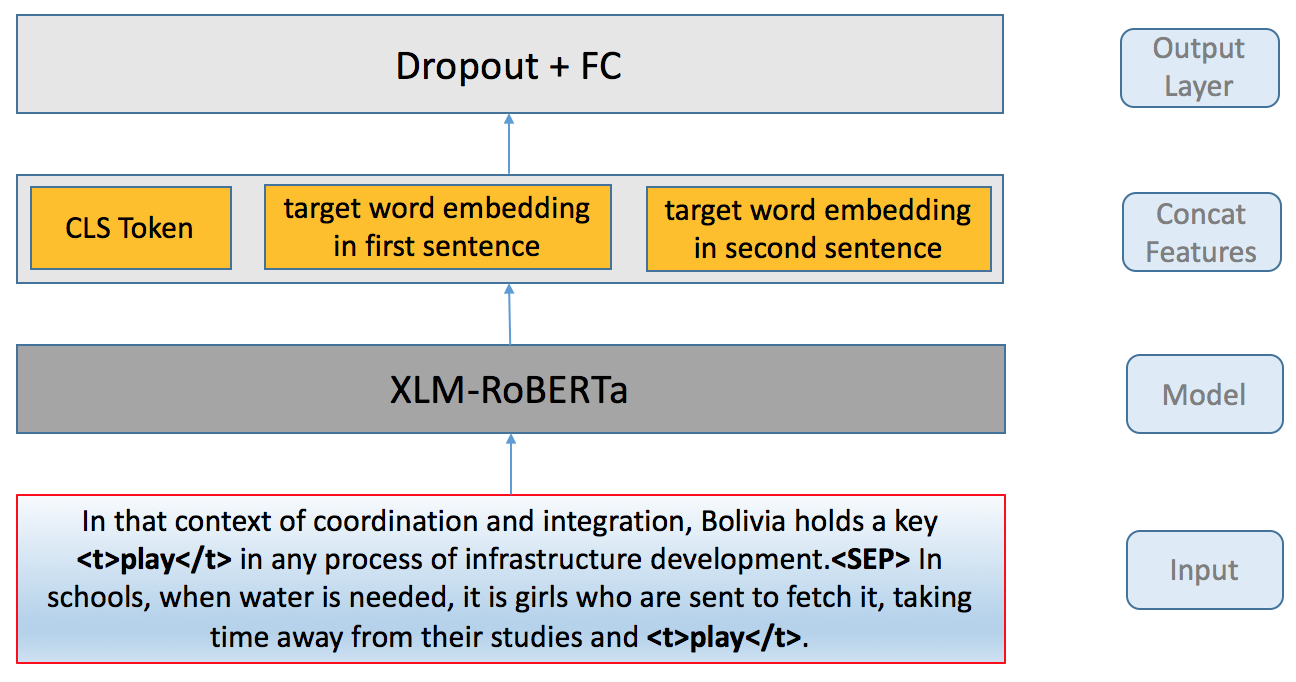}
\caption{Fine-tuned XLM-RoBERTa model architecture. \label{fig:framework}}
\end{figure*}
The training dataset consists of two files in the JSON format: the .data file and the .gold file.  The .data file contains the following information: unique id of the pair, target lemma, part-of-speech in \{NOUN, VERB, ADJ, ADV\}, the first sentence, the second sentence, the start and the end indices (zero-based numbering) of the target word in the first and the second context, respectively.  The .gold file contains unique id of the pair and the label, which is represented by T or F. 

For the training dataset, we clean up the text by completing word abbreviation, removing special punctuation, and segmenting the sentences into subword lists by Byte-Pair Encoding (BPE)~\cite{BPE2015}.  Since it is difficult to capture the meaning of the target word in the context for long sentences~\cite{pan2019recent,zhu2021retrieving}, we limit the length of each sentence with maximum 40 words before and after the target word.

We include additional resource, WordNet, to augment our training data because WordNet is a large lexical database of English.  Nouns, verbs, adjectives and adverbs are grouped into sets of cognitive synonyms (synsets), each expressing a distinct concepts.  Here, we randomly select example sentences of target word in WordNet to expand our training corpus, which increases around 30\% of training data.  By such preprocessing, we obtain the dataset and report the statistics in Table~\ref{tab:data_stat}.

\subsection{Model Design}
\Cref{fig:framework} outlines our model architecture, which consists of four modules, i.e., input design, model learning, final feature construction, and the classifier.  The whole framework is based on fine-tuning the pre-trained XLM-RoBERTa model to conduct binary classification on two given contexts.  Different from the inputs for XLM-RoBERTa, the input of our system contains of the following modifications: first, in order to highlight the target word in the contexts, we borrow the setting in~\cite{lei2017swim,DBLP:conf/emnlp/BothaSG20} by adding special symbols $<$t$>$ and $<$/t$>$ to embrace the target word in the contexts.  Given the example presented in \cref{fig:ex}, the target word of ``play" is then embraced by the additional symbols, $<$t$>$ and $<$/t$>$ in the contexts.  Second, we concatenate the given two contexts by $<$SEP$>$.  \Cref{fig:framework} illustrates the result in the input module.  Moreover, in the experiment, we exchange the order of the contexts to get more training data. 

After learning the tokens' representations by XLM-RoBERTa, we construct a new vector by concatenating the [CLS] token's representation in the last layer of XLM-RoBERTa and the representations of the target word in both sentences.  As BPE tokenization may separate a target word into several subwords, we compute its representation by averaging the corresponding representations.  Next, the newly constructed feature is fed into a fully-connected network to compute the final binary prediction probability.  

\begin{table*}[htbp]
    \centering
    \begin{tabular}{l|@{~}c@{~}|@{~}c@{~}|@{~}c@{~}|@{~}c@{~}|@{~}c@{~}|@{~}c@{~}|@{~}c@{~}|@{~}c@{~}|@{~}c@{~}|@{~}c@{~}}
    \hline
   Strategy & Avg & En-En & Fr-Fr & Ru-Ru & Zh-Zh & Ar-Ar & En-Ru & En-Zh & En-Fr & En-Ar\\\hline
   {{Base}}& 80.8 & 85.5 & 80.7 &78.7 &80.9 &79.1 &81.0 &81.9 &79.1 &80.4\\
   \hline
   \multirow{1}{*}{{Large}}& 85.1 & 88.2 & 84.2& 84.3&87.0 &82.6 &84.6 &85.7 &85.6&83.9\\
   \hline
   \multirow{1}{*}{{Large + RO}}& 85.4 &88.7 &85.3 &85.1 &86.9 &83.3 &84.7 &85.9 &84.7 &83.6\\
   \hline
   \multirow{1}{*}{{Large + RO +  LRA}}& 85.4 & 89.2 &84.9 &85.1 &86.8 &83.2 &84.8 &85.6 &85.2 &84.1\\\hline 
   {{Large + RO + LRA + ES}}& 85.5 &89.0 &84.8 &85.5 &86.9 &83.1 &85.2 &85.2 &85.4 &84.1\\  
   \hline
   \multirow{1}{*}{{Large + RO + CTWE}}& 86.3 & 90.0 &85.8 &85.9 &87.1 &83.9 &86.0 &86.1 &85.4 &86.2\\
   \hline
   \multirow{1}{*}{{Large + RO + CTWE + HC}}& 86.3 &89.9 &85.7 &85.9 &87.2 &84.2 &85.6 &86.7 &85.0 &86.3\\ 
   \hline
   {{Large + RO + CTWE + HC}} &\multirow{2}{*}{86.5}& \multirow{2}{*}{91.6} & \multirow{2}{*}{85.8} & \multirow{2}{*}{85.7} & \multirow{2}{*}{87.1} & \multirow{2}{*}{84.0} & \multirow{2}{*}{85.3} & \multirow{2}{*}{87.1} & \multirow{2}{*}{85.6} & \multirow{2}{*}{86.0}\\  
   + WordNet& & & & & & & & & & 
   \\   \hline
   {{Large + RO + CTWE + HC }}& \multirow{2}{*}{87.0} & \multirow{2}{*}{91.1} & \multirow{2}{*}{86.3} & \multirow{2}{*}{85.9} & \multirow{2}{*}{87.9} & \multirow{2}{*}{85.1} & \multirow{2}{*}{86.3} & \multirow{2}{*}{87.2} & \multirow{2}{*}{86.3} & \multirow{2}{*}{86.9}\\ 
   + WordNet + AT && & & & & & & & &  \\
   \hline
   {{Large + RO + CTWE + HC}}& \multirow{2}{*}{\bf 88.1} & \multirow{2}{*}{\bf 91.7} & \multirow{2}{*}{\bf  86.9} & \multirow{2}{*}{\bf 86.5} & \multirow{2}{*}{\bf 89.2} & \multirow{2}{*}{\bf 86.5} & \multirow{2}{*}{\bf 88.0} & \multirow{2}{*}{\bf 87.9} & \multirow{2}{*}{\bf 88.6} & \multirow{2}{*}{\bf 87.2}\\ 
   + WordNet + AT + PL & & & & & & & & & & \\
  \hline
    \end{tabular}
    \caption{Results of fine-tuning XLM-RoBERTa under different strategies.  The abbreviation is defined as follows: Base: XML-RoBERTa$_{\tiny\mbox{Base}}$; Large: XML-RoBERTa$_{\tiny\mbox{Large}}$; RO: Ranger Optimizer; LRA: learning rate adjustment; ES: early stop; CTWE: concatenating target words' embeddings; HC: the best parameters for LRA and ES; AT: adversarial training; PL: pseudo labels. \label{tab:result}}   
\end{table*}

During training, we conduct the following techniques to increase the model convergence and robustness: 
\begin{compactitem}
\item {\bf Optimizer.}  We adopt the Ranger~\cite{DBLP:conf/eccv/YongHHZ20} optimizer to replace the AdamW because it is a more synergistic optimizer combining rectified Adam and the look-ahead mechanism with gradient centralization in one optimizer.   
\item {\bf Adversarial training.} We apply the fast gradient method~\cite{DBLP:conf/iclr/MiyatoDG17} in the training to obtain more stable word representations.
\item {\bf Cross validation.}  We also apply stratified $K$-fold cross validation on the training set and the development set.  For each fold, we hold the group as a local test set and set the remaining groups as the training set.  We then average the model prediction on each fold as the final prediction to obtain more robust results.   
\item {\bf Pseudo labelling.} Pseudo labelling \cite{DBLP:journals/tip/WuP18} is an effective semi-supervised learning method to utilize the abundant unlabeled data via their pseudo labels.  In this work, we first train our model on the training set.  Next, we apply the trained model to predict the En-En multi-lingual test set and use the predicted labels as our pseudo labels.  Finally, both the training set and the En-En pseudo labels are included to train a final model.  Especially, we observe that by this trick, this final model can improve the prediction performance on cross-lingual tasks slightly.
\end{compactitem}
It is noted that in the cross-lingual tasks, we do not back-translate the subwords to English but apply the same model trained from the En-En dataset because it allows us to maintain the target word in the corresponding languages seamlessly.  This is similar to the procedure in~\cite{DBLP:conf/acl/ArtetxeRY20}.

\section{Experiments}
In the following, we detail our experimental setup and present the results with analysis.

\subsection{Setup}
Our code is written in Pytorch based on the Huggingface Transformer library\,\footnote{\url{https://github.com/huggingface/transformers}} for XLM-RoBERTa.  Other hyperparameters are set based on our hand-on experience.  For example, the seed for the random generator is set to 3,999.  The batch size is set to 10 and the hidden feature size is 1,024.  The maximum length limit of a context is 240 though it is unreachable because we have conducted trimming in the data preprocessing procedure.   The dropout rate is tested from \{0.2, 0.3, 0.25, 0.28\} and finally fixed to 0.28.  $K$ in the stratified cross validation is set to 5. The two special tokens, $<$t$>$ and $<$/t$>$, are included into the word dictionary for learning. 

The training data consists of the official En-En multilingual training corpus and the contexts from WordNet.  At the beginning, we choose XLM-RoBERTa$_{\tiny\mbox{Base}}$ as the backbone of our system to explore the possibility of our implementation tricks.  After identifying the effectiveness of the designed input in Sec.~3.2, we apply XLM-RoBERTa$_{\tiny\mbox{Large}}$ to tune the corresponding hyperparameters, such as changing the learning rate, the batch size, the dropout rate, and the early stop mechanism.  Furthermore, we observe that long contexts may ignore the importance of the target word in the contexts.  Hence, we center on the target words to cut off the contexts at both ends with a certain length.  To further strengthen the influence of the target word in a context, we concatenate the embedding of the [CLS] token with the embeddings of the target word in the contexts as the final input for the logit fully-connected network.   From our experiment, this strategy can significantly boost the model performance while improving the convergence.  

\subsection{Results}
Table 2 reports the results of different implementation strategies on the tasks.  From the results, we observe that 

\begin{compactitem}
\item By replacing XLM-RoBERTa$_{\tiny\mbox{Base}}$ with XLM-RoBERTa$_{\tiny\mbox{Large}}$, we can gain at least 3\% improvement on all tasks.
\item By applying Ranger optimizer, we attain the results in  Large+RO, which gain an average increase of 0.2\% per task.  We conjecture the improvement comes from the fact that the model converges to a more optimal solution. 
\item In Large+RO+LRA, we vary the learning rate from 1.5e-5, 1.3e-5, 1.2e-5, to 1.21e-5 progressively and finally find that when the learning rate is 1.2e-5, we attain the best performance.  We then search the optimal epoch for the early stop by setting the maximum number of epoch to 10. In Large+RO+LRA+ES, we observe the optimal epoch for early stop (the patience value) is 3.  These parameters are then fixed for HC.  From the results, we notice that tuning the learning rate and adopting the early stop mechanism can improve the model performance accordingly. 
\item By concatenating target the word embedding, we obtain the results in Large+RO+CTWE, and actually, our model can be trained with fewer epochs and attain around 1.1\% improvement on average.
\item By adding more training data from WordNet, we get another 0.2\% average improvement in Large+RO+CTWE+HC+WordNet.  We conjecture the improvement mainly comes from the increase of the training data.
\item By adding the Pseudo label data, we can gain another 0.8\% average improvement. The score of EN-EN test dataset is generally higher than other test dataset.  We discover that the first 462 pieces of English test dataset have the same target word as test dataset in other tasks. Therefore, adding EN-EN pseudo label helps predict other tasks.
\end{compactitem}

In sum, we conclude that by applying  XLM-RoBERTa$_{\tiny\mbox{Large}}$ on the Ranger optimizer, the target word embedding concatenation mechanism, more external training data, and pseudo labels, we can improve the model performance accordingly.

Finally, our system attains the champion on the En-Ar, En-Fr, En-Ru, and En-Zh cross-lingual tasks.  In multilingual tasks, we also sit at eighth place, seventh place, sixth place, seventh place, and fifth place for the En-En, Ar-Ar, Fr-Fr, Ru-Ru, Zh-Zh tasks, respectively.

\section{Conclusion}
In this paper, we present our system to tackle the word-in-context disambiguation task.  We fine-tune the  XLM-RoBERTa model to solve both multilingual and cross-lingual word-in-context disambiguation tasks.  We specifically design the input format to emphasize the target word in two contexts and promote the importance of the target word by concatenating the embeddings in the corresponding context with the [CLS] token to output the classification probability.  We apply several training tricks to improve the robustness of model and attain improvement during this procedure.  The competition results demonstrate the effectiveness of our implementation.  In the future, we plan to explore more model architecture to boost the performance for multilingual tasks.

\bibliographystyle{acl_natbib}
\bibliography{anthology,acl2021}

\end{document}